\documentclass[runningheads]{llncs}

\usepackage{eccv}

\usepackage{eccvabbrv}

\usepackage{graphicx}
\usepackage{booktabs}
\usepackage[numbers]{natbib}
\usepackage{siunitx}
\usepackage[skip=0pt, indent=20pt]{parskip}
\usepackage[accsupp]{axessibility}  

\usepackage[pagebackref,breaklinks,colorlinks,citecolor=eccvblue]{hyperref}

\usepackage{orcidlink}

\usepackage{xcolor}

\begin{document}
\title{Scaling Up Resonate-and-Fire Networks for Fast Deep Learning}

\author{Thomas E. Huber\inst{1,2}\orcidlink{0009-0004-6894-3061} \and
Jules Lecomte\inst{1}\orcidlink{0000-0002-7103-0843} \and
Borislav Polovnikov\inst{3}\orcidlink{0009-0001-7295-9232} \and
Axel von Arnim\inst{1}\orcidlink{0000-0002-9680-7591}}

\authorrunning{Huber et al.}

\institute{Fortiss GmbH, Guerickestraße 25, 80805 München, Germany \and
Technische Universität München, Arcisstraße 21, 80333 München, Germany \and
Ludwig-Maximilians-Universität, Geschwister-Scholl-Platz 1, 80539 München, Germany}

\maketitle

\begin{abstract}
  Spiking neural networks (SNNs) present a promising computing paradigm for neuromorphic processing of event-based sensor data. The resonate-and-fire (RF) neuron, in particular, appeals through its biological plausibility, complex dynamics, yet computational simplicity. Despite theoretically predicted benefits, challenges in parameter initialization and efficient learning inhibited the implementation of RF networks, constraining their use to a single layer. 
  In this paper, we address these shortcomings by deriving the RF neuron as a structured state space model (SSM) from the HiPPO framework.
  We introduce \textit{S5-RF}, a new SSM layer comprised of RF neurons based on the S5 model, that features a generic initialization scheme and fast training within a deep architecture.  
  S5-RF scales for the first time a RF network to a deep SNN with up to four layers and achieves with $78.8\%$ a new state-of-the-art result for recurrent SNNs on the Spiking Speech Commands dataset in under three hours of training time. 
  Moreover, compared to the reference SNNs that solve our benchmarking tasks, it achieves similar performance with much fewer spiking operations. Our code
is publicly available at \url{https://github.com/ThomasEHuber/s5-rf}.

\end{abstract}

\section{Introduction}

\label{sec:intro}
Neuromorphic computing is a set of principles applied to software and hardware with the goal of developing power and data efficient computing paradigms that can compete with traditional artificial neural networks (ANNs). Neuromorphic event-based vision, which relies on visual sensors capturing asynchronous per-pixel brightness changes in a scene, is a promising aspect of this domain and offers perspectives characterized by higher spatio-temporal resolutions trade-off, high dynamic range, power efficiency, and asynchronous sparse data output. Notably, such sensors can be combined with Spiking Neural Networks (SNNs) for an end-to-end neuromorphic pipeline~\cite{Pfeiffer2018Oct,Lu2022Aug}. 

The main building block of SNNs is based on a biologically inspired model of neuronal computation through time-dependent spike trains. In this approach, a neuron is modeled by a characteristic ordinary differential equation (ODE) that governs the temporal evolution of the neuron's internal state, combined with a non-linear spiking mechanism responsible for its firing output. A spiking neuron intrinsically deals with sequential data and presents a sequence-to-sequence map, where its internal dynamics can be understood as a solution to the defining ODE. The most common implementation is given by the notorious leaky integrate-and-fire (LIF) neuron~\cite{Abbott1999Nov,Lu2022Aug}, but biologically more plausible models such as the resonate-and-fire (RF) neuron are slowly gaining popularity~\cite{Izhikevich2001,Auge2021Sep,Frady2022Oct,Lehmann2023May}. Notably, both neurons feature internal linear dynamics.
At the same time, studies of RF-SNNs have, so far, been limited to rudimentary exploitation of the RF's resonating behavior, such that a pure resonator network on par with traditional ANNs and SNNs has remained elusive.

Recently, \emph{structured state space models} (SSMs) have been receiving growing interest as general purpose sequence-to-sequence models \cite{Gu-S4, Gu-S4D, smith-S5}.
SSMs have been developed to model long-range dependencies in sequential data by leveraging the \textit{high-order polynomial projection operators} (HiPPO) framework \cite{Gu-HiPPO, Gu-How2TrainHiPPO}.
Their basic idea is to map a sequence's history onto an orthonormal basis in an online fashion, resulting in a linear ODE which describes the time dynamics of the sequence's basis coefficients.
The initial computational algorithm proposed by S4 \cite{Gu-S4} has favorable time complexity, but is non-trivial.
By approximating the S4-ODE as a diagonalized ODE, S4D \cite{Gu-S4D} and S5 \cite{smith-S5} could match similar performance as S4 with equal time complexity by utilizing convolutions in time and parallel scans, respectively. \par 
In this work, we show how RF-SNNs can be reinterpreted as deep SSMs based on the HiPPO framework \cite{Gu-HiPPO}, allowing for an implementation of RF neurons that can be trained through the back-propagation through time algorithm (BPTT)~\cite{neftci2019surrogate}. Building upon the SSM architecture S5, we establish a link between SSMs and RF-SNNs and propose a new architecture, \textit{S5-RF}. 
With this layer, we are able to scale RF neurons to deep SNNs, achieving state-of-the-art performance for recurrent SNNs on the Spiking Speech Commands (SSC) dataset with a training time of under three hours.
Moreover, compared to current state-of-the-art SNNs, S5-RF achieves similar accuracy with fewer parameters and fewer spiking operations on the sequential benchmarks sMNIST, psMNIST, as well as on the Spiking Heidelberg Dataset (SHD) and SSC.

After presenting related literature in \cref{sec:RelatedWork}, we proceed by introducing the Izhikevich RF neuron~\cite{Izhikevich2001} from the point of view of its internal ODE dynamics and then presenting a minimal theory of the S5 layer in \cref{sec:RF} and \cref{sec:SSM}. 
In \cref{sec:model}, we introduce the S5-RF layer by deriving the RF dynamics from a SSM.
We report benchmark results of S5-RF in terms of accuracy and computational efficiency in \cref{sec:experiments} and \ref{sec:results}, followed by an ablation on our design choices in \cref{sec:ablation}.
Finally, we summarize the results of this work in \cref{sec:conclusion}.

\section{Related Work}
\label{sec:RelatedWork}
The non-linearity of the spiking dynamics and the time dimension inherent to SNNs make their training harder than for conventional ANNs. Nevertheless, the advent of different techniques such as the use of surrogate gradients and BPTT \cite{neftci2019surrogate}, or conversion of already trained ANNs to SNNs \cite{ANN2SNN}, have enabled to train SNNs and made them competitive with ANNs. Recent works in SNNs have shown increased accuracies, encoding spike delays and using neurons as coincidence detectors \cite{hammouamri2023learning,Patiño-Saucedo23delays}. 

In particular RF neurons have gained a recent interest with their incorporation into the available set of implemented hardware neurons of the Loihi 2 chip from Intel \cite{orchard2021efficientneuromorphicsignalprocessing}. Besides use cases for rudimentary spike encoding~\cite{Lehmann2023May,Auge2021Sep} and short-time Fourier transform (STFT)~\cite{Frady2022Oct}, RF networks have successfully been trained with BPTT~\cite{nmnistRAF} on the neuromorphic MNIST dataset \cite{10.3389/fnins.2015.00437}. Additionally, a recent work introduced custom, modified resonator dynamics to train a stable resonator network on the more complex task of keyword spotting~\cite{higuchi2024balanced}.

SSMs have successfully been adopted into SNNs by incorporating LIF neurons as activations. In particular the LMUFormer \cite{liu2024lmuformer} builds upon the Legendre Memory Unit \cite{voelker2019legendre} for speech recognition, while a spiking S4 model \cite{du2024spiking} has been employed for speech enhancement. Finally, a spiking stochastic S4 layer termed S6 \cite{bal2024rethinking} builds upon the LIF neuron by expanding the scalar hidden state to a multi-dimensional hidden state and by adding stochasticity into the spiking mechanism.

\section{Resonate-and-Fire Neurons}
\label{sec:RF}
We take the point of view that biological neurons are essentially dynamical systems which are characterized by specific, generally non-linear ODEs~\cite{izhikevich2007dynamical}. The famous Hodgkin-Huxley (HH) model~\cite{Hodgkin1952}, for example, builds upon the biological insight on the different ionic channels across a neuron's membrane and is widely considered to be one of the most biologically accurate spiking neuron models~\cite{Izhikevich2004Sep}.

Resonate-and-fire neurons were introduced by Izhikevich in 2001~\cite{Izhikevich2001} as the simplest mathematical model that can exhibit persistent subthreshold oscillations as observed in the HH model \cite{Izhikevich2001}. A key feature of the RF neuron is its linear internal dynamics, which significantly reduces computational costs while retaining more of the HH's biological plausibility than the simplest LIF neuron. The RF model is formulated as an ODE for the complex-valued function $z(t)$,
\begin{equation}
\label{eq:RF}
\frac{d}{dt} z(t) = (-b + i\omega) z(t) + I(t) \, ,
\end{equation}
where $b > 0$ is a decay parameter, $\omega \in \mathbb{R}$ the resonance frequency of the neuron, and $I(t)$ is the real-valued input current. Mathematically, this is equivalent to the damped harmonic oscillator problem, extended by a spike-generating mechanism. Contrary to the internal dynamics, the spike-generating function is generally a non-linear function of both the amplitude and the phase of $z$. A typical condition that we use in this work is that a spike is fired whenever the real part $\Re{(z)}$ surpasses some threshold. 

It is important to emphasize that since Eq.~\eqref{eq:RF} is linear, we can write down an analytical solution that takes the form
\begin{equation}\label{eq:RF_analytic}
    z(t) = e^{(-b + i\omega)(t-t_0)} z(t_0) + \int_{t_0}^t e^{(-b+i\omega)(t-\tau)} I(\tau) d\tau \, .
\end{equation}
%
This is equivalent to a STFT of the input signal~\cite{Frady2022Oct} and ultimately corresponds to a linear basis transformation of the time-dependent input. 
 
Due to this last property, the RF neuron has been employed in LIF networks as a spike encoding for continuous data~\cite{Auge2021Sep,Lehmann2023May}. 
In the remainder of this work, we derive a RF neuron formulation as a SSMs and arrive at a generic layer architecture, S5-RF. This enables the scaling of RF layers to deep networks and places RF-SNNs on par with the best of traditional LIF-SNNs.

\section{Structured State Space Models}
\label{sec:SSM}
In the following, we proceed by introducing SSMs as deep neural networks with a focus on the recently introduced S5 model. After defining and characterizing SSMs through a set of necessary properties, we show how they are practically implemented in a discretized way, and illustrate how to construct them based upon the theoretical HiPPO framework.

\subsection{Definition}
In most general terms, linear time-invariant SSMs are sequential models defined as dynamical systems that map an input signal $u(t) \in \mathbb{C}^N$ through a hidden state $x(t) \in \mathbb{C}^H$ to an output signal $y(t) \in \mathbb{C}^M$,
\begin{align}
        \label{eq:ssm}
        \frac{d}{dt} x(t) &= Ax(t)+Bu(t) \, , \\
        \label{eq:ssm2}
        y(t) 			 &= Cx(t) + Du(t) \, ,
\end{align}
with the state matrix $A \in \mathbb{C}^{H \times H}$, the input matrix $B \in \mathbb{C}^{H \times N}$, the output matrix $C \in \mathbb{C}^{M \times H}$, and the skip matrix $D \in \mathbb{C}^{M \times N}$ as the learnable parameters.

Typically, sequential models capture the history of an input signal through a hidden state which forms the main approach to long-range modeling (compare, e.g., with the LSTM \cite{Hochreiter-LSTM} or the GRU \cite{cho-GRU} architectures). In SSMs, this process of \textit{memorization} is governed by the state matrix $A$. In the case when $A = V \Lambda V^{-1}$ is diagonalizable with a diagonal $\text{diag}(\lambda_1, \ldots, \lambda_H)=\Lambda \in \mathbb{C}^H$ and some matrix $V \in \mathbb{C}^{H \times H}$, then transforming the hidden state according to $\tilde{x} = V^{-1} x$ brings Eqs. (\ref{eq:ssm},~\ref{eq:ssm2}) into a diagonal form as well:
\begin{align}
        \label{eq:diag_ssm}
        \frac{d}{dt}\tilde{x}(t) &= \Lambda \tilde{x}(t)+V^{-1}Bu(t) \, , \\
        \label{eq:diag_ssm2}
        y(t) &= CV \tilde{x}(t) + Du(t) \, .
\end{align}
Here, the different components of the hidden state become disentangled and follow mutually independent dynamics determined by the corresponding eigenvalues of $A$. Notably, for any given SSM, the transformation matrix $V$ can be precomputed upon initialization, leading to a long-range behavior controlled by $O(H)$ rather than $O(H^2)$ parameters.

We further emphasize that the dynamical system defined by Eqs. (\ref{eq:diag_ssm},~\ref{eq:diag_ssm2}) is only stable if all components of $\Lambda$ have a negative real part \cite{dorf2017}. In the process of learning, however, this condition could potentially be broken. A common strategy to mitigate this problem is by initializing all $\Re(\lambda_i)$, $i=1, \ldots, H$, as their log-values and exponentiating during the forward pass \cite{goel2022s} -- a step that we found crucial for the model performance in this work.

\subsection{Discretization}
The model described by Eqs. (\ref{eq:ssm},~\ref{eq:ssm2}) or (\ref{eq:diag_ssm},~\ref{eq:diag_ssm2}) is formulated as a continuous time process, so any practical implementation requires a suitable discretization scheme. The most common approaches, e.g. forward Euler methods or zero-order hold (ZOH) \cite{Pechlivanidou2022Jun}, involve a manual choice of a time step $\Delta \in \mathbb{R}$ that should be picked as small as possible in order to minimize the local and global truncation errors. In ZOH, for example, the discretized form of Eqs. (\ref{eq:ssm},~\ref{eq:ssm2}) becomes
%
\begin{align}
        \label{eq:linear_recurrence}
		x_{k} &= \bar{A}x_{k-1} + \bar{B}u_k \, , \\
        \label{eq:linear_recurrence2}
        y_k &= \bar{C}x_k + \bar{D}u_{k} \, ,
\end{align}
%
with the discretized matrices 
\begin{equation}
\label{eq:ZOH_matrices}
    \bar{A} = \exp(\Delta A), \quad \bar{B} = (A)^{-1}\left(\exp(\Delta A) - I\right) B, \quad \bar{C} = C, \quad \bar{D} = D \, .
\end{equation}
In event-like data such as spike trains in SNNs that come with a natural time step, the truncation error can be avoided altogether, allowing for full simulation of the SSM's true dynamics. The key ingredient is to represent a discrete spike train in continuous time as a weighted Dirac comb,
\begin{equation}
    u(t) = \sum_{n=1}^{N} \delta(t-t_n)u_n \, ,
\end{equation}
where $t_n$ denotes the time of the $n$-th event and $u_n$ the corresponding weighting. With this input signal, the SSM in Eqs. (\ref{eq:ssm},~\ref{eq:ssm2}) with the initial condition $x(t_0) = 0$ allows for the analytic solution
\begin{equation}
    x(t) = \int_{t_0}^{t} \exp{\left(A (t-\tau)\right)} B u(\tau) d\tau = \sum_{\substack{ n=1 \\ t_n \leq t}}^{N} \exp{\left(A(t-t_n)\right)} Bu_n \, ,
\end{equation}
%
which can be rewritten as discrete recurrence giving the Dirac discretization
%
    \begin{align}
        x_k &= \sum_{n=1}^{k} \exp{\left(A(t_k-t_n)\right)} Bu_n  \\
        &= \exp{\left(A(t_k-t_{k-1})\right)} \left(\sum_{n=1}^{k-1} \exp{\left(A(t_{k-1}-t_n)\right)} Bu_n\right) + Bu_k  \\
        \label{eq:dirac-disc}
        &= \exp{\left(\Delta A\right)} x_{k-1} + Bu_k \, .
    \end{align}
%
Here, $\Delta = t_k-t_{k-1}$ is the intrinsic time step in the equidistant input data, which allows to apply Eq.~\eqref{eq:dirac-disc} irrespective of whether an event is present at a given time or not. Finally, we note that using the above discretization scheme for the diagonalized model in Eq. (\ref{eq:diag_ssm},~\ref{eq:diag_ssm2}) further simplifies the simulated dynamics since the exponential of $\exp{\left(\Delta \Lambda\right)}$ reduces to $H$ individual scalar exponents.

\subsection{HiPPO-Initialization}
Historically, the initialization of the matrix $A$ has been the limiting factor for the performance of SSMs. This is because $A$ is crucial for governing the dynamics of the hidden state, influencing the long-range process of memorization. This limitation has only been lifted once the HiPPO theory~\cite{Gu-HiPPO, Gu-How2TrainHiPPO} introduced a robust initialization scheme based on real-time function approximation. The first key insight was based on interpreting the hidden state $x(t)$ as a basis representation of the truncated input signal $u(s)|_{s\leq t}$ with respect to a suitably chosen, time-dependent basis of a $H$-dimensional subspace of $L^2((-\infty,t])$.
The second key insight was to differentiate $x$ through time which resulted in a SSM, allowing for online learning without having to recalculate $x$ for each timestep $t$. 

\begin{figure}[t]
    \centering
    \includegraphics[width=\textwidth]{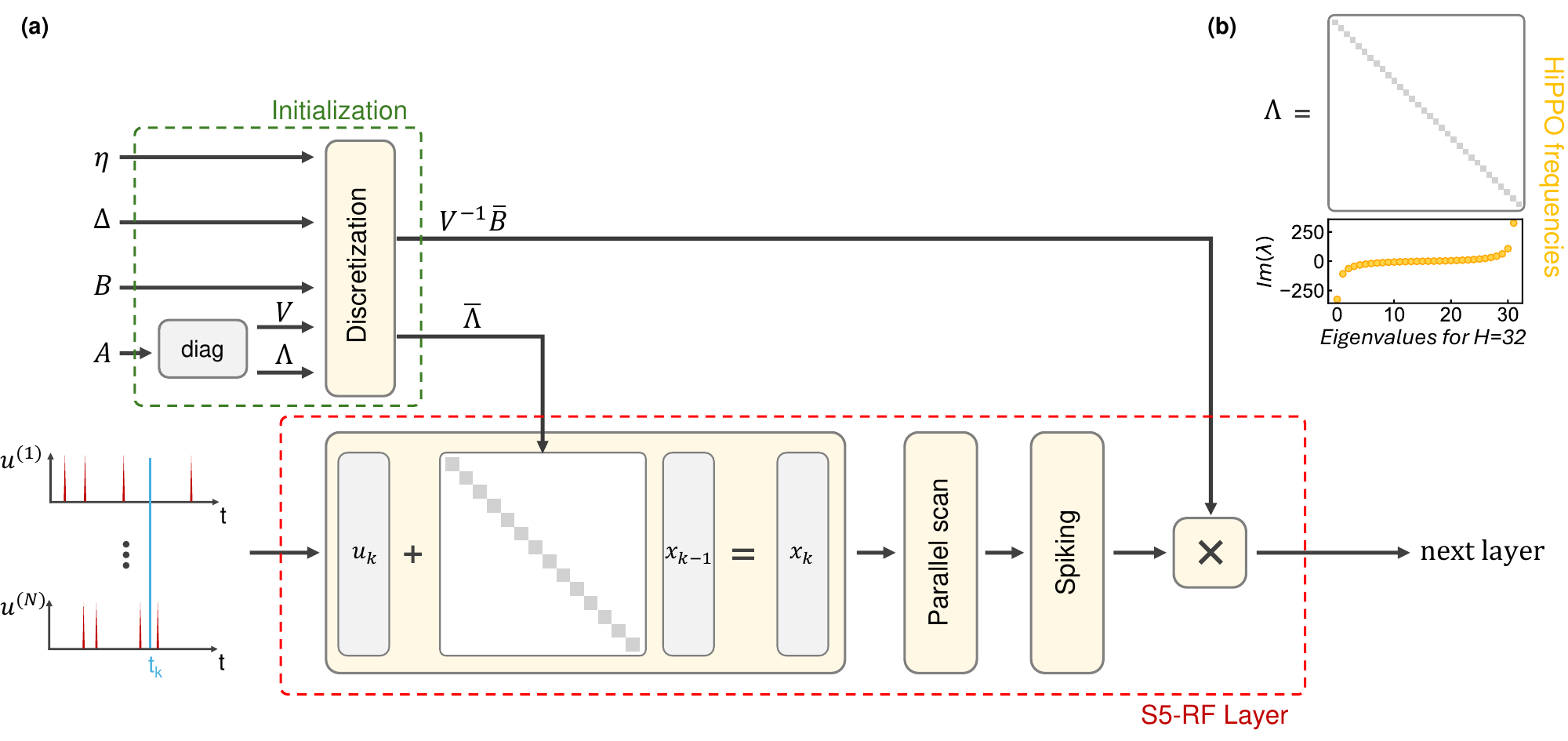}
    \caption{(a) S5-RF layer architecture based upon the S5 architecture~\cite{smith-S5}. (b) Imaginary values of the components of $\Lambda_{N}$ in the case of $H=32$ dimensions. We note that since the matrix $A_{N}$ is real, its eigenvalues necessarily come in complex-conjugated pairs.}
    \label{fig1}
\end{figure}

Based on HiPPO, the most common initialization is built on the system of scaled Legendre polynomials and is derived from an exponentially decaying measure that defines the inner product on $L^2((-\infty,t])$~\cite{Gu-HiPPO,Gu-How2TrainHiPPO}. The resulting matrix $A_{LegS}$ \cite{Gu-S4, Gu-S4D, smith-S5} leads to optimal compression of the input's history, allowing for long-range modeling of sequential data~\cite{Gu-HiPPO}. Unfortunately, this comes at the cost of considerable computational complexity that arises from repeated matrix multiplication in the process of learning. A straightforward solution would be to diagonalize $A_{LegS}$, yet this is numerically unstable due to exponentially growing eigenvalues~\cite{Gu-S4}. Luckily, it has been shown that $A_{LegS} = A_{N} + pp^\dagger$ can be decomposed as a normal-plus-low-rank matrix 
%
%
where $A_{N}$ approximates $A_{LegS}$ in the limit of large state spaces~\cite{Gu-S4D}, i.e. the low-rank part $pp^\dagger$ becomes negligible. Importantly, $A_{N}$ is stably diagonalizable and all of its eigenvalues have a real part of $-0.5$ \cite{Gu-S4D}. The imaginary parts are shown in \cref{fig1}b. We emphasize that a non-zero imaginary part leads to oscillatory behavior as mentioned in \cref{sec:RF}, so anticipating the relation to RF neurons, we choose to call them HiPPO frequencies.

The S5 layer defines a SSM with a block-diagonal state transition matrix where each block is initialized with $A_{N}$.
It is diagonalized with a similar matrix formed from the eigenvectors $V_N$ of $A_{N}$. 
S5 solves the recurrence from Eq.~\eqref{eq:linear_recurrence} as associative parallel scans leveraging prefix sums \cite{blelloch1990prefix}. 
This allows for efficient parallelization, requiring a minimum of $O(\log(L))$ operations, where $L$ denotes the sequence length, given sufficiently many parallel processors as described above.


We emphasize that classical sequential models such as LSTM or GRU, or current SNNs based on LIF or ALIF~\cite{yin2021accurate}, include non-linearities inside of the sequential processing. This renders them unparallelizable along the sequence length and poses a significant computational bottleneck that limits the model in hidden size, depth, and applicability to large data sets.




\section{Our Approach: S5-RF}
\label{sec:model}

Ultimately, both RF neurons and SSMs are essentially represented by linear ODEs.
As no satisfying implementations of RF neurons that allow for deep SNNs have been found, we argue that the RF neuron should be understood as a SSM and therefore initialized and discretized according to the HiPPO framework.
In this section, we first represent the RF neuron in the context of the S5 SSM before developing a proper Dirac discretization scheme based on HiPPO.
Finally, we discuss the spiking mechanism and its impact on computational efficiency.

\subsection{From SSM to Deep RF}
To see the similarity between SSMs and RF neurons, we start by reconsidering a discrete SSM initialized with $A_{N}$ and omit the skip matrix. Upon diagonalization, this system yields
    \begin{align}
        \frac{d}{dt}x &= \Lambda_{N}x + V_{N}^{-1}Bu \, ,\\ 
        y& = CV_{N}x \, .
    \end{align}
Adding a non-linear spike generating function $\Theta: \mathbb{C} \rightarrow \{1, 0\}$ after the $x$ dynamics results in the dynamical system
\begin{align}
        \frac{d}{dt}x &= \Lambda_{N}x + V_{N}^{-1}Bu \, , \\  
        s &= \Theta(V_{N}x) \, ,\\
        y& = Cs \, ,
\end{align}
which is already reminiscent of the RF neuron. At this point, the only difference is given by the $V_{N}$ matrix in the spike-generating function and the two matrices $V_{N}^{-1}B$ and $C$ on the input and output channels. In deep networks, the output of one layer is simultaneously the input of the next one, so in practice, the output and input matrices are multiplied to form one connection matrix, $V_{N}^{-1}BC$. Additionally, we note that both $B$ and $C$ are randomly initialized learnable synaptic weights, therefore we can ultimately consider $BC$ as one randomly initialized connection matrix, noted $B$.

As far as the $V_{N}$ in the spiking mechanism is concerned, we note that, as a basis transformation, $V_{N}$ is necessarily a unitary matrix. A unitary transformation $U$, however, always maps a HiPPO-SSM onto a HiPPO-SSM as long as the matrices $A$ and $B$ are mapped to $U^{-1}AU$ and $U^{-1}B$, respectively~\cite{Gu-How2TrainHiPPO}. We can therefore reconsider the diagonalization of Eqs. (\ref{eq:ssm},~\ref{eq:ssm2}) and simply apply the unitary transformation to the matrices $A$ and $B$, while dropping $V_{N}$ from the spike generating function.

The resulting SSM is equivalent to a layer of RF neurons and takes the form 
%
\begin{align}
        \frac{d}{dt} {x} &= \Lambda_{N}x + V_{N}^{-1}Bu \, , \\    
        \label{eq:RF_SSM}
        s &= \Theta(x) \, , \\
        y& = s \, .
\end{align}
%
Here, the real and imaginary parts of $\Lambda_{N}$ form the neurons' decay and frequency, respectively, while the complex learnable weights $V_{N}^{-1}B$ present the synaptic connections. In light of the derivation based on the S5 layer, we choose to call this RF neuron \textit{S5-RF} and show its model architecture in \cref{fig1}a. Notably, due to the chosen initialization we no longer consider a single RF neuron performing a STFT, but rather a collection of neurons modeling the evolution of the sequence's history within the HiPPO framework. The immediate advantage of this generic initialization scheme is that it is both task and data independent.

\subsection{Dirac-SSM Discretization}
Both in the S5 architecture as well as in other structured SSMs like the S4 and the S4D layers, the discretization step $\Delta$ has two different, yet conflicting interpretations. On the one hand, it is the step size between consecutive time steps which indicates that it should be fixed during training. On the other hand, it is a scaling factor that modulates the integration measure defining the inner product on $L^2 ((-\infty,t])$, thereby providing a mechanism to assign different weights to past time steps~\cite{Gu-S4, Gu-S4D}.
The latter property is, therefore, responsible for the time range of memorization of each individual neuron, making it beneficial to learn $\Delta$ as well. 
Throughout the literature, $\Delta$ has been motivated by the former but deployed following the latter interpretation.

Notably, the confusion arising from these two different perspectives is closely related to the choice of discretization. As it has been shown in Gu \etal~\cite{Gu-How2TrainHiPPO}, multiplying the matrices $A$ and $B$ of a HiPPO-SSM by a scalar modulates the underlying measure as described above. This applies to the ZOH discretization, where both $A$ and $B$ are multiplied by $\Delta$.
For Dirac discretization (Eq. \eqref{eq:dirac-disc}), however, this is no longer the case. For this reason, in this work, we first explicitly multiply $A$ and $B$ by a learnable scalar $\eta > 0$ and then discretize the SSM with a fixed $\Delta$ that plays the role of a constant time step. ZOH results in
\begin{equation}
    \label{eq:our-zoh-disc}
    \bar{A} = \exp(\eta \Delta A ), \quad 
    \bar{B} = (\eta \Delta A)^{-1}(\bar{A} - I) \eta \Delta B \\, \quad 
    \bar{C} = C, \quad 
    \bar{D} = D \, ,
\end{equation}
whereas with Dirac discretization, the discrete matrices become
\begin{equation}
    \label{eq:our-dirac-disc}
    \bar{A} = \exp(\eta \Delta A), \quad \bar{B} = \eta B, \quad \bar{C} = C, \quad \bar{D} = D .
\end{equation}

We use this Dirac-SSM discretization for any intermediate and the input layer if the input is an event stream.
Otherwise, we discretize the input layer with ZOH.

\subsection{Spiking without Losing Speed}
\label{sec:spike}

Traditionally, models of spiking neurons include a refractory period that ensures a hard reset to the internal dynamics once the neuron has fired -- a mechanism used to prevent continuous bursts of spikes in order to ensure sparsity and energy efficiency. 
However, with a hard reset, the associative property of the internal neuron's dynamics is broken which is necessary to deploy prefix sums \cite{blelloch1990prefix} like in S5. For this reason, we do not add a refractory period after spiking.  

We let a neuron spike whenever the real part is larger than some threshold. 
Since the neuron's internal dynamics are characterized by a decaying oscillation, we rely on it to stop spiking.
The spiking mechanism can be written in terms of a Heaviside function $H$

%
\begin{equation}
    \Theta (x_k) = H(\Re(x_k) - \xi)) \, ,
\end{equation}
with the threshold value $\xi$ that we set to $1$. This formulation allows us to use the surrogate gradient method~\cite{neftci2019surrogate} and train all parameters with BPTT.

\section{Experiments}
\label{sec:experiments}
We report the performance of S5-RF networks in terms of accuracy and computational efficiency averaged over five random seeds per benchmark. Our architecture consists of multiple S5-RF layers of the same size. To encode the input dimension to the neuron layer size, we use a learnable linear projection. As our decoder, we connect the last S5-RF layer to leaky-integrate neurons with a learnable time constant. This step is necessary to avoid linking the class prediction with the frequency of a single RF neuron. Afterward, we apply mean pooling over the sequence length and calculate the cross entropy loss.
Skip connections are added to combat vanishing gradients whenever layer dimensions permit.

As the surrogate gradient method, we choose the multi Gaussian \cite{yin2021accurate} which is also used by ALIF. Similar to S5, we use a higher learning rate for the neuron connections than for the neuron parameters and a cosine annealing scheduler \cite{loshchilov2016sgdr}. 
Finally, we Adam~\cite{kingma2014adam} as our optimizer and apply weight decay only to the inter-neuron connections.

To leverage efficient parallelization, our model is implemented with JAX~\cite{jax2018github} and the Equinox~\cite{kidger2021equinox} deep learning framework. All of our experiments were conducted on a single RTX 4060 Ti GPU.

\subsection{Sequential MNIST}
First, we evaluate the S5-RF on sequential MNIST (sMNIST) and its permuted version (psMNIST) which are benchmarks with reported spiking behavior.
On sMNIST, the $28$ by $28$ pixel greyscale image of handwritten digits is converted to a pixel-by-pixel stream, forming a sequence of length $784$.
In psMNIST, the pixels are additionally randomly permuted. 
Both datasets contain $60,000$ samples of which $10,000$ are reserved for testing. 
We split $10\%$ off from the train set, to create a validation set.

In both datasets, the sequence is not event-based, so following the above discussion, we discretize the first layer with ZOH. In this case, we also include the $V_N$ matrix in the spike-generating function, but keep it fixed throughout training. In the subsequent layers, we apply the SSM-Dirac discretization and drop the $V_N$ matrix as in Eq.~\eqref{eq:RF_SSM}.
With this benchmark, we show that our model can be adapted to non-neuromorphic data via the discretization method.  

\subsection{Audio Processing}
We also evaluate our model on relevant neuromorphic benchmarks including the \textit{Spiking Heidelberg Dataset} \cite{Cramer_2022} (SHD) and the \textit{Spiking Speech Command} (SSC). Both datasets contain audio recordings of spoken words that are converted into an event stream by an artificial cochlea model, and we use the tonic library~\cite{lenz_gregor_2021_5079802} to load both datasets.

The SHD consists of approximately $10\,000$ recordings of spoken digits in English and German.
About $2\,000$ recordings are reserved for the test set with two speakers unseen during training.
For a fair comparison, we use the test set as the validation set, similar to the procedure in the literature.
The SSC is a spiking variant of the Google Speech Command Dataset \cite{warden2018speech} with $35$ different English words.
It is about ten times larger than the SHD, with $70\%$ making up the train set, $10\%$ the validation set, and $20\%$ the test set, although the conditions are less controlled since all speakers are present in every set.

Before training, we bin the audio recordings to a length of $250$ steps, with each bin forming an event if there is at least one spike in the bin.
We further downsample the input channels from $700$ to $140$. 
This significantly reduces the number of parameters of the linear encoder, which would have otherwise made up the bulk of the network.
To help generalization, we randomly shift $20\%$ of the time all channels up or down.
Similar to Event-SSM~\cite{schone2024scalable}, we use an event-based cutmix~\cite{yun2019cutmix} version to prevent overfitting on SSC. 
Before feeding a batch to the model, we randomly sample a time interval, cut it out, and replace it with the same time interval from a different recording in the same batch. 
The labels are fused based on the spike ratio of previous and inserted spikes.

\section{Results}
\label{sec:results}
In \cref{tab:SHD-SSC}, we compare the performance of S5-RF with current SNN baselines in terms of test accuracy, model size, neuron type, and whether they are recurrent on SSC and SHD.
S5-RF achieves an accuracy of $91.86 \%$ on SHD and $78.8 \%$ on SSC, outperforming RadLIF on SSC, the current state-of-the-art method in recurrent SNNs, with much fewer parameters.
It is only surpassed by the recently proposed DCLS-Delays~ \cite{hammouamri2023learning} and Event-SSM.
While the former is using convolutions, the latter is not a SNN.
During our experiments, we noticed that the event-based cutmix and channel shifting were crucial to prevent overfitting, whereas methods like channel jitter were insufficient. 
Furthermore, with four layers on SSC, the S5-RF can be effectively scaled to a deep SNN.

\begin{table}[t]
\centering
\caption{Results of SHD and SSC benchmarks based on neuron type (LIF, RF, SRM for Spike Response Model, or N/a for not applicable), whether they are recurrent, number of parameters, and accuracy.}\label{tab:SHD-SSC}
\begin{tabular}{clcccc}
\hline
Dataset & Model & Neuron Type & Recurrent & \#Params & Accuracy \\ 
\hline
SHD & EventProp-GeNN \cite{nowotny2024lossshapingenhancesexact} & LIF & Yes & - & $84.80\%$ \\
& Cuba-LIF \cite{Dampfhoffer22CubaLIF} & LIF & Yes & $0.14$M & $87.80\%$ \\
& Adaptiv SRNN \cite{yin2021accurate} & LIF & Yes & - & $90.40\%$ \\
& BRF \cite{higuchi2024balanced} & RF & Yes & $0.1$M & $90.4\%$ \\
& SNN+Delays \cite{Patiño-Saucedo23delays} & LIF & No & $0.1$M & $90.43\%$ \\
& TA-SNN \cite{Yao21TemporalAttention} & LIF & No & - & $91.08\%$ \\
& STSC-SNN \cite{Yu2022ConvAttention} & LIF & No & $2.1$M & $92.36\%$ \\
& Adaptive Delays \cite{Sun23AdaptivAxonalDelays} & SRM & No &$0.1$M & $92.45\%$ \\
& DL128-SNN-Dloss \cite{Sun2023LearnableAxonalDelays} & SRM & No & $0.14$M & $92.56\%$ \\
& BHRF \cite{higuchi2024balanced} & RF & Yes & $0.1$M & $92.7\%$ \\ 
& RadLIF \cite{Bittar2022SurrogatGradientBaseline} & LIF & Yes & $3.9$M & $94.62\%$ \\
& DCLS-Delays \cite{hammouamri2023learning} & LIF & No & $0.2$M & $95.07\%$ \\
& Event-SSM \cite{schone2024scalable} & N/a & Yes & $0.4$M & $95.45\%$ \\
& \textbf{S5-RF (ours)} & \textbf{RF} & \textbf{Yes} & $\bold{0.07}$\textbf{M} & $\bold{90.65\%}$ \\
& \textbf{S5-RF (ours)} & \textbf{RF} & \textbf{Yes} & $\bold{0.2}$\textbf{M} & $\bold{91.86\%}$ \\
\hline
SSC & Recurrent SNN  \cite{cramer2020heidelberg} & LIF & Yes & - & $50.90\%$ \\
& Heter. RSNN \cite{perez2021neural} & LIF & Yes &- & $57.30\%$ \\
& SNN-CNN \cite{sadovsky2023speech} & LIF & No &- & $72.03\%$ \\
& Adaptive SRNN \cite{yin2021accurate} & LIF & Yes & - & $74.20\%$ \\
& SpikGRU \cite{Dampfhoffer22CubaLIF} & LIF & Yes & $0.28$M & $77.00\%$ \\
& RadLIF \cite{Bittar2022SurrogatGradientBaseline} & LIF & Yes & $3.9$M & $77.40\%$ \\
& DCLS-Delays \cite{hammouamri2023learning} & LIF & No & $0.7$M & $79.77\%$ \\
& DCLS-Delays \cite{hammouamri2023learning} & LIF & No & $2.5$M & $80.69\%$ \\
& Event-SSM \cite{schone2024scalable} & N/a & Yes & $0.6$M & $87.1\%$ \\
& \textbf{S5-RF (ours)} & \textbf{RF} & \textbf{Yes} & $\bold{0.7}$\textbf{M} & $\bold{76.85\%}$ \\
& \textbf{S5-RF (ours) }& \textbf{RF} & \textbf{Yes} & $\bold{1.7}$\textbf{M} & $\bold{78.8\%}$ \\
\hline
\end{tabular}
\end{table}

\begin{figure}
    \centering
    \includegraphics[width=0.9\linewidth]{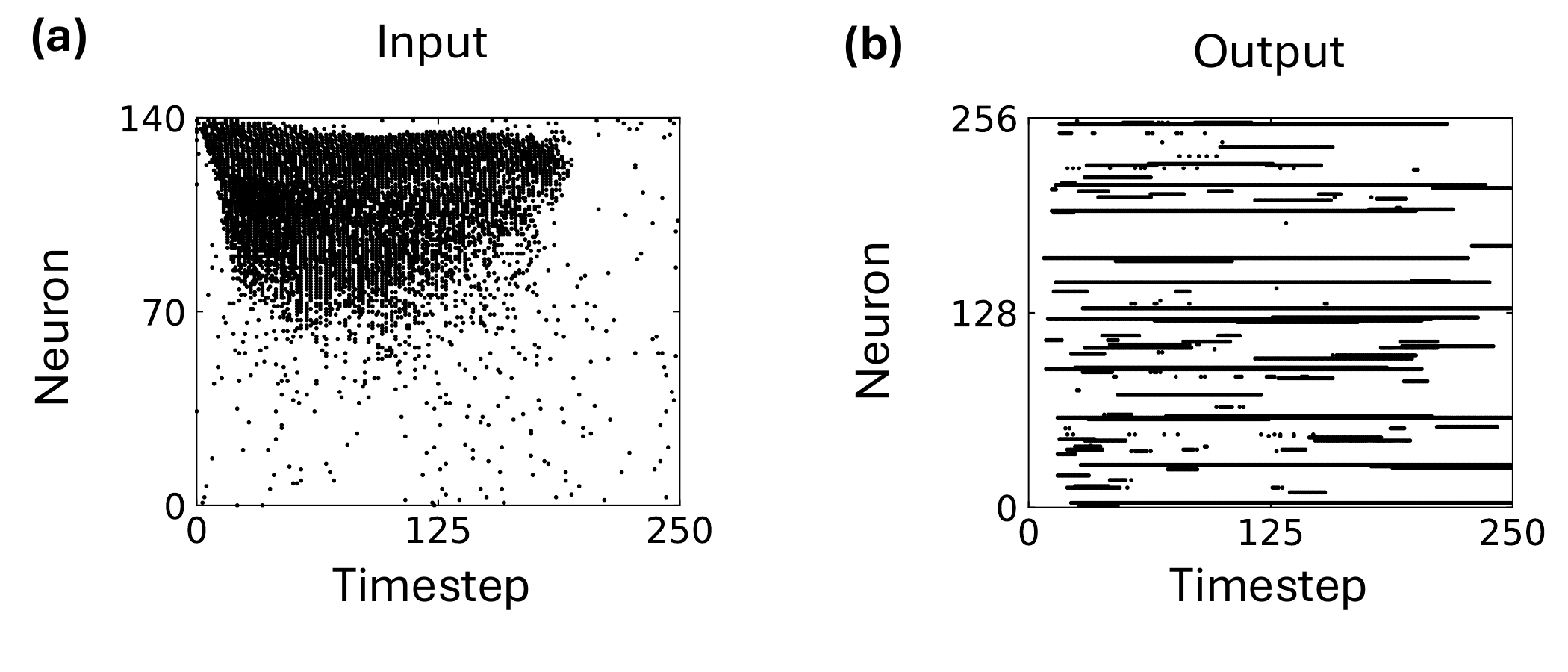}
    \caption{\textbf{Spiking activity of in- and output of a S5-RF layer trained on SHD.} 
    (a) Sample input of the word "nine" with locally dense spiking activity across neurons. (b) Output of a layer with some neurons bursting and others remaining silent.} 
    \label{fig:3}
\end{figure}

We also compare the computational efficiency of the S5-RF on the average number of spiking operations (SOP) and train time with other methods from the literature, see \cref{tab:SRNNs}.
Throughout all benchmark datasets, S5-RF performs similarly to its competitors in terms of accuracy, but with fewer parameters and a $10$ to $20$ times shorter train time.
Most notably, it took under three hours to train a four layer network on SSC.
We want to emphasize that this would not have been possible with a refractory period as discussed in ~\cref{sec:spike}.
Despite the absence of refractory periods, the S5-RF's spiking behavior is sparse with up to three and a half times fewer SOPs than its competitors. Even with more layers and more neurons per layer, S5-RF requires fewer SOPs than ALIF on SHD.  \Cref{fig:3} depicts a bursting behavior for a small fraction of neurons, while others are never activated, showing the high selectivity and biological plausibility of our model.
In contrast to BRF or BHRF \cite{higuchi2024balanced}, S5-RF does not require any engineered conditions restricting the neuron dynamics to achieve this behavior which prohibited BRF and BHRF from being scaled up.
The S5-RF outperforms ALIF \cite{yin2021accurate} and does not require a spike encoding as a preprocessing step.

\begin{table}[]
\centering
\caption{
    Results of the SRNN architectures comparing their efficiency based on the number of parameters, parallelizability, accuracy, average number of spiking operations (SOPs), and train time (ALIF from Yin \etal~\cite{yin2021accurate} except for train time, BRF, BHRF, and ALIF train time from Higuchi \etal~ \cite{higuchi2024balanced}).
}\label{tab:SRNNs}
\begin{tabular}{clcccccc}
\hline
Dataset        & Model         & Architecture        & \#Params & Parallelizable & Acc        & SOPs    & Train Time   \\\hline
sMNIST  & ALIF    & (1,64,256x2,10)  & - & No & $98.7\%$ & $70\, 811$  & $>\SI{30}{\hour}$             \\ 
        & BRF           & (1,256,10)          & $68\,874$ & No & $99.0\%$ & $15\, 463$ & $\SIrange{20}{30}{\hour}$       \\
        & BHRF          & (1,256,10)          & $68\,874$ & No & \textbf{$99.1\%$} & $21\, 566$ & $\SIrange{20}{30}{\hour}$      \\
        & \textbf{S5-RF}         & \textbf{(1,128x2,10)}      & \textbf{36\,362} & \textbf{Yes} & $\bold{98.89 \%}$ & $\bold{15\,547}$ & \textbf{$\SI{1.5}{\hour}$}                \\ 
\hline
psMNIST & ALIF    & (4,64,256x2,10)  & - & No & $94.3\%$ & $59\,772$  & $>\SI{30}{\hour}$             \\
        & BRF           & (1,256,10)          & $68\,874$ & No  & $95.0\%$ & $27\,840$ & $\SIrange{20}{30}{\hour}$      \\
        & BHRF          & (1,256,10)          & $68\,874$ & No  & $95.2\%$ & $24\,564$ & $\SIrange{20}{30}{\hour}$      \\
        & \textbf{S5-RF} & \textbf{(1,128x2,10)} & \textbf{36\,362} & \textbf{Yes} & $\bold{95.29\%}$ & $\bold{16\,062}$ & \textbf{$\SI{1.5}{\hour}$} \\
\hline
SHD     & ALIF   & (700,128x2,20)    & - & No & $90.4\%$ & $24\,690$ & - \\
        & BRF           & (700,128,20)      & $108\,820$ & No & $90.4\%$ & $4\,692$ & - \\   
        & BHRF          & (700,128,20)      & $108\,820$ & No & $92.7\%$ & $4\,140$ & - \\
        & \textbf{S5-RF} & \textbf{(140,128x2,20)} & \textbf{74\,516} & \textbf{Yes} & $\bold{90.65\%}$ & $\bold{6\,944}$ & \textbf{$\SI{7.5}{\min}$} \\
        & \textbf{S5-RF} & \textbf{(140, 256x2,20)} & \textbf{214\,548} & \textbf{Yes} & $\bold{91,86\%}$ & $\bold{8\,431}$ & \textbf{$\SI{7.8}{\min}$} \\
\hline
SSC     & ALIF    & (700,400x2,35)    & - & No & $74.2\%$ & $19\,450$ & - \\
        & \textbf{S5-RF} & \textbf{(140,512x2,35)} & \textbf{706\,595} & \textbf{Yes} & $\bold{76.85\%}$ & $\bold{11\,904}$ & \textbf{$\SI{1.2}{\hour}$}  \\
        & \textbf{S5-RF} & \textbf{(140,512x4,35)} & \textbf{1\,758\,243} & \textbf{Yes} & $\bold{78.8\%}$ & $\bold{21\,751}$ & \textbf{$\SI{2.5}{\hour}$} \\
 \hline
\end{tabular}
\end{table}

\section{Ablation}
\label{sec:ablation}
To assess the relative roles of the S5-RF constituents shown in \cref{fig1}, we ablate the initialization with the HiPPO frequencies as well as the SSM-Dirac discretization for HiPPO on the SSC dataset. For the former, we follow Higuchi \etal~\cite{higuchi2024balanced} by randomly initializing the decay $b \sim U(2, 3)$ and the frequency $\omega \sim U(5, 10)$. For the latter, we fix $\eta = 1$ during the training with random initialization. Notably, we stabilize the neuron by enforcing the decay to stay positive during training. Finally, we drop the eigenvectors $V_N$ from the equations because they are meaningless under random initialization.

\begin{figure}[t]
    \centering
    \includegraphics[width=0.9\textwidth]{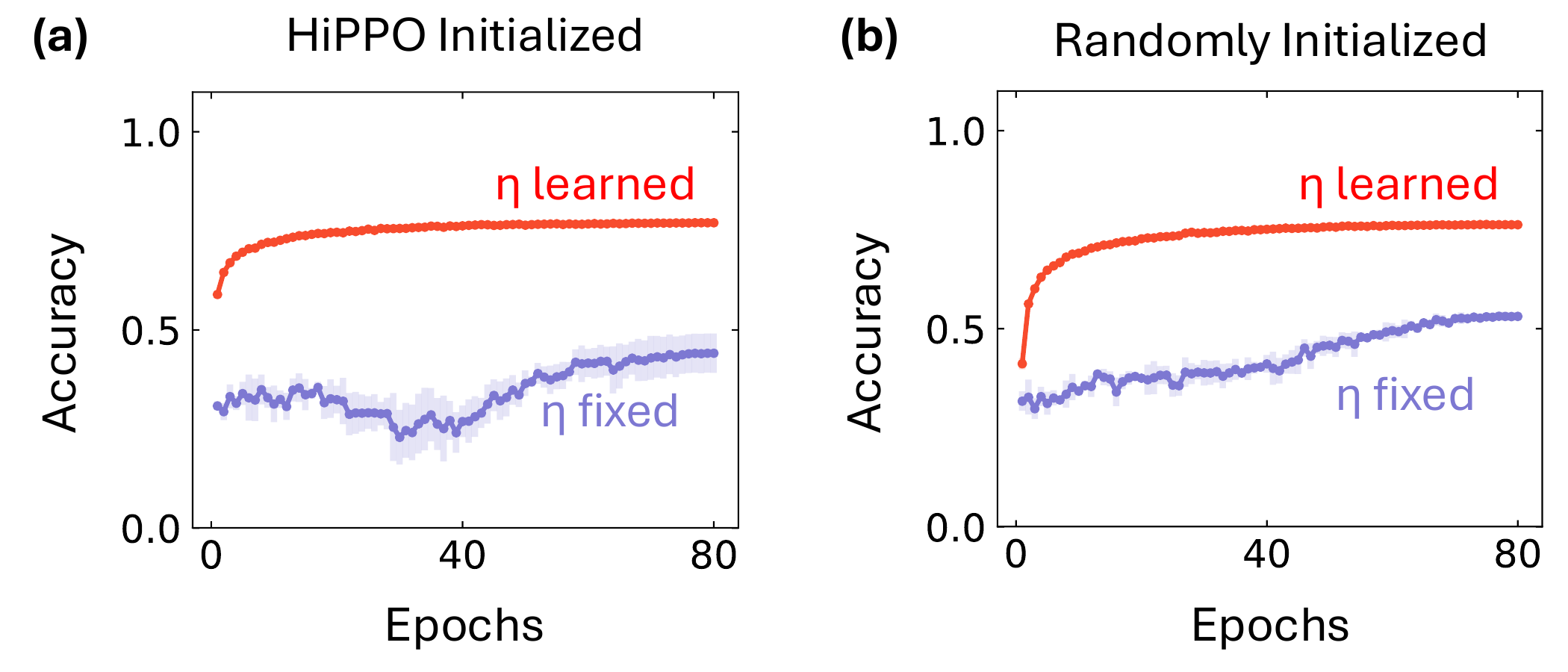}
    \caption{\textbf{Ablation results in a 512x2 S5-RF network.} (a) Test accuracy for the HiPPO initialization with fixed (bottom line, blue) and learned (top line, red) $\eta$, respectively. The shaded area around the curves represents the standard deviation obtained from five different random seeds. (b) Test accuracy for random initialization.}
    \label{fig2}
\end{figure}

\Cref{fig2} shows the ablation results with a two layers S5-RF network, consisting of $512$ neurons per layer with a block size of $32$.
We emphasize that fixing $\eta$ leads to a significant drop in performance on both the HiPPO as well as the random initialization, highlighting the necessity of learning the memorization range. 
When $\eta$ is learned, on the other hand, S5-RF presents remarkable robustness to the initialization conditions where both initializations lead to efficient training, with the random scheme matching the HiPPO frequencies. We note that such surprising behavior has not yet been observed in the SSM literature, which assumes the HiPPO initialization to be a necessary condition for effective learning~\cite{Gu-S4, Gu-S4D, smith-S5}. 
In that context, our results show that the initialization of the RF-SSM is secondary to the much more critical Dirac-SSM discretization with a learnable $\eta$. While a full investigation of these results is outside of the scope of this paper, we hypothesize that interpreting the HiPPO theory as a variant of a STFT $-$ thus establishing a direct correspondence between HiPPO and RF neurons $-$ would shed further light on our observations. 

\section{Conclusion}
\label{sec:conclusion}
In this work, we provide a new understanding of the RF neuron through the lens of SSMs. 
We theoretically derive the RF neuron from the HiPPO framework, allowing us to interpret the neuron's hidden state as a basis decomposition and to find a general initialization scheme of the RF neuron independent of the dataset at hand.
We show empirically that our proposed S5-RF neuron scales effectively to deep SNNs, achieving state-of-the-art performance on SSC for recurrent SNNs. Our proposed model is computationally efficient in time and convinces with its low number of parameters and sparsity of spikes. Additionally, the biological plausibility of the Izhikevich RF neuron is preserved by only addressing the discretization and initialization schemes.
Finally, our ablation study shows the necessity of the Dirac-SSM discretization scheme as well as the robustness of S5-RF to different initialization conditions.

To the best of our knowledge, this is the first resonator network, that has been successfully scaled to a deep SNN with up to four layers.  
Notably, we believe our method can be further improved by using the event timing in the discretization, or by applying learnable synaptic delays to the spiking mechanism.

\textbf{Acknowledgements.} This work was initiated at the 2023 Munich neuroTUM hackathon with the contribution of Reem al Fata and Eric Armbruster.



%
%
\bibliographystyle{splncs04}
\bibliography{ref}
\end{document}